\documentclass[runningheads,a4paper]{llncs}

\usepackage{amssymb}
\usepackage{graphicx}
\usepackage{times}

\usepackage{comment} 
\usepackage{graphicx} 
\usepackage{epstopdf} 
\usepackage{amsmath} 
\usepackage{mathtools} 
\usepackage{mathrsfs} 
\usepackage{scrextend} 
\usepackage{url} 
\usepackage{epstopdf}
\usepackage{bm} 
\usepackage{amssymb} 
\usepackage{color}
\usepackage{flushend} 

\makeatletter
\def\hlinew#1{%
	\noalign{\ifnum0=`}\fi\hrule \@height #1 \futurelet
	\reserved@a\@xhline}
\makeatother

\newcommand{\tabincell}[2]{\begin{tabular}{@{}#1@{}}#2\end{tabular}} 
\usepackage{url}
\newcommand{\keywords}[1]{\par\addvspace\baselineskip
\noindent\keywordname\enspace\ignorespaces#1}

\begin{document}

\mainmatter  

\title{Deep Multi-instance Networks with Sparse Label Assignment for Whole Mammogram Classification}

\titlerunning{Deep MIL with Sparse Label Assignment for Whole Mamm Class.}

%
%
\author{Wentao Zhu, Qi Lou, Yeeleng Scott Vang, and Xiaohui Xie}
%
\authorrunning{W. Zhu et al.}

\institute{University of California, Irvine}
\institute{\{wentaoz1, xhx\}@ics.uci.edu, \{qlou, ysvang\}@uci.edu}

%
%

\toctitle{Lecture Notes in Computer Science}
\tocauthor{Authors' Instructions}
\maketitle

\begin{abstract}
  Mammogram classification is directly related to computer-aided diagnosis of breast cancer. Traditional methods requires great effort to annotate the training data by costly manual labeling and specialized computational models to detect these annotations during test. Inspired by the success of using deep convolutional features for natural image analysis and multi-instance learning for labeling a set of instances/patches, we propose end-to-end trained deep multi-instance networks for mass classification based on whole mammogram without the aforementioned costly need to annotate the training data. We explore three different schemes to construct deep multi-instance networks for whole mammogram classification. Experimental results on the INbreast dataset demonstrate the robustness of proposed deep networks compared to previous work using segmentation and detection annotations in the training. 
\end{abstract}

\section{Introduction}\label{sec:intro}
According to the American Cancer Society, breast cancer is the most frequently diagnosed solid cancer and the second leading cause of cancer death among U.S. women. Mammogram screening has been demonstrated to be an effective way for early detection and diagnosis, which can significantly decrease breast cancer mortality~\cite{oeffinger2015breast}. However, screenings are usually associated with high false positive rates, high variability among different clinicians, and over-diagnosis of insignificant lesions~\cite{oeffinger2015breast}. To address these issues, it is important to develop fully automated robust mammographic image analysis tools that can increase detection rate and meanwhile reduce false positives.

Traditional mammogram classification requires extra annotations such as bounding box for detection or mask ground truth for segmentation. These methods rely on hand-crafted features from mass region followed by classifiers~\cite{varela2006use}. The main barrier to use hand-crafted features is the associated cost of time and effort. Besides, these features have potential poor transferability for use in other problem settings because they are not data driven.
Other works have employed different deep networks to detect region of interest (ROI)
and obtained mass boundaries in different stages \cite{dhungel2016automated}.
However, these methods require training data to be annotated with bounding boxes and segmentation ground truths which require expert domain knowledge and costly effort to obtain.

Due to the high cost of annotation, we intend to perform classification based on a raw, un-annotated whole mammogram. Each patch of a mammogram can be treated as an instance and a whole mammogram is treated as a bag of instances. The whole mammogram classification problem can then be thought of as a standard multi-instance learning problem.
Thus, we propose three different schemes, i.e., max pooling, label assignment, and sparsity, to perform deep multi-instance learning for the whole mammogram classification task.

\begin{figure}[t]
	\begin{center}
		\begin{minipage}{\linewidth}
			\centerline{\includegraphics[width=\textwidth]{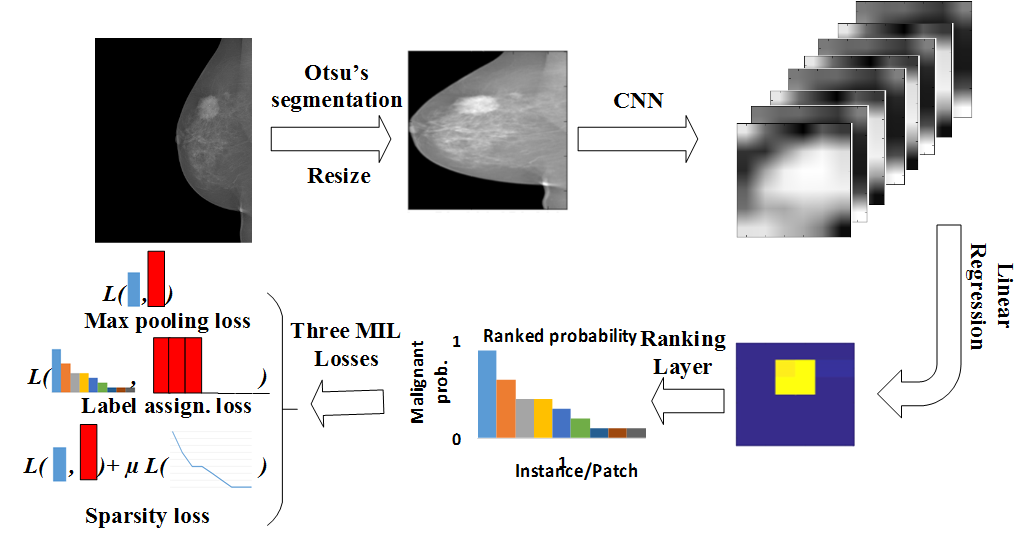}}
		\end{minipage}
		\caption{The proposed deep multi-instance network framework. First, we use Otsu's segmentation to remove the background and resize the mammogram to $224\times224$. Second, the deep multi-instance network accepts the resized mammogram as input to the convolutional layers. Third, linear regression with weight sharing is employed for the malignant probability of each position from the convolutional neural network (CNN) feature maps of high channel dimensions. Then the responses of the instances/patches are ranked. Lastly, the learning loss is calculated using max pooling loss, label assignment, or sparsity loss for the three different schemes.}
		\label{fig:framework}
	\end{center}
\end{figure}

\begin{figure}[t]
	\begin{center}
		\begin{minipage}{0.245\linewidth}
			\centerline{\includegraphics[width=\linewidth]{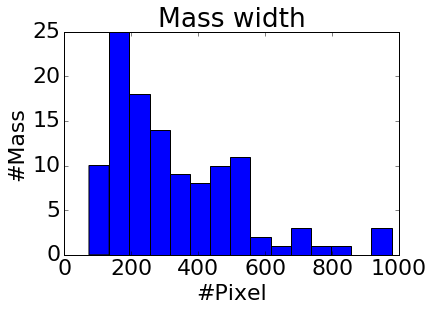}}
			\center{(a)}
		\end{minipage}
		\begin{minipage}{0.245\linewidth}
			\centerline{\includegraphics[width=\linewidth]{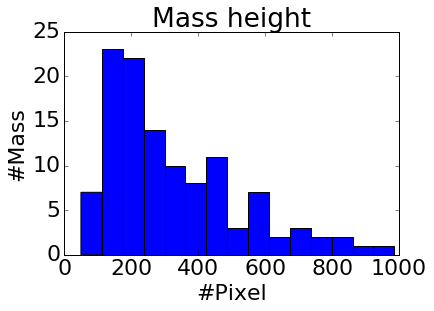}}
			\center{(b)}
		\end{minipage}
		\begin{minipage}{0.245\linewidth}
			\centerline{\includegraphics[width=\linewidth]{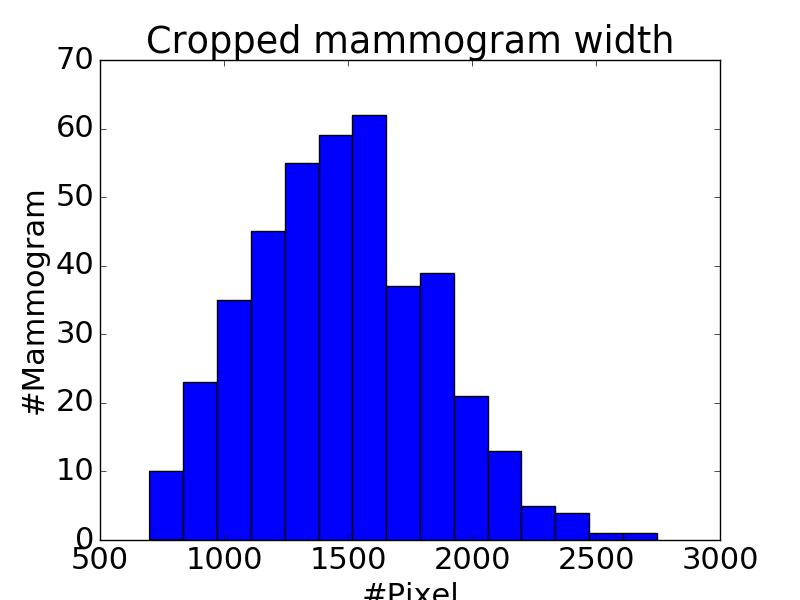}}
			\center{(c)}
		\end{minipage}
		\begin{minipage}{0.245\linewidth}
			\centerline{\includegraphics[width=\linewidth]{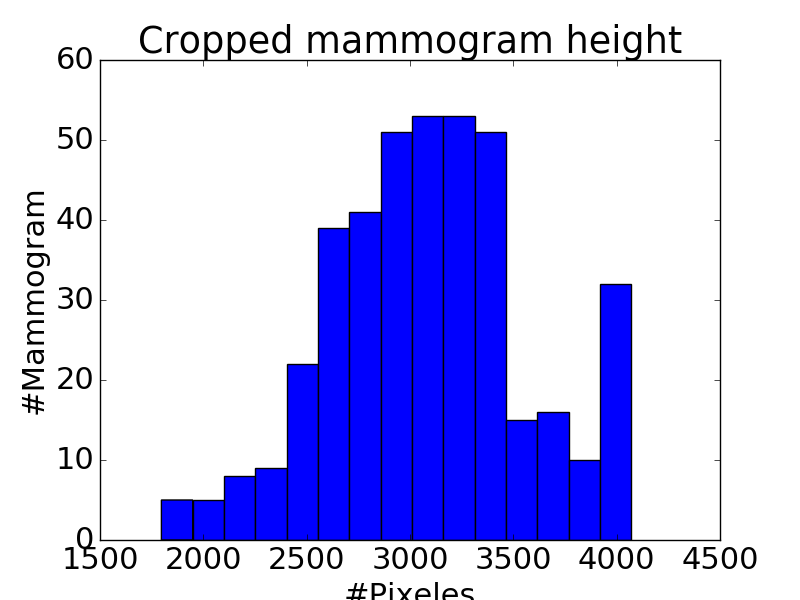}}
			\center{(d)}
		\end{minipage}
		\caption{Histograms of mass width (a) and height (b), mammogram width (c) and height (d). Compared to the size of whole mammogram ($1,474 \times 3,086$ on average after cropping), the mass of average size ($329 \times 325$) is tiny, and takes about 2\% of a whole mammogram. }
		\label{fig:mass}
	\end{center}
\end{figure}

The framework for our proposed end-to-end deep multi-instance networks for mammogram classification is shown in Fig.~\ref{fig:framework}. To fully explore the power of deep multi-instance network, we convert the traditional multi-instance learning assumption into a label assignment problem. Specifically, we also propose a more efficient, label assignment based deep multi-instance network. As a mass typically composes only 2\% of a whole mammogram (see Fig.~\ref{fig:mass}), we further propose sparse deep multi-instance network which is a compromise between max pooling-based and label assignment-based multi-instance networks. The proposed deep multi-instance networks are shown to provide robust performance for whole mammogram classification on the INbreast dataset~\cite{moreira2012inbreast}.

\section{Related Work}\label{sec:rel}
\subsection{Mammogram Classification}\label{sec:mammdl}
Beura et al. designed co-occurrence features and used wavelet transform for breast cancer detection~\cite{beura2015mammogram}.
Several works have used deep networks to perform mammogram mass classification~\cite{jiao2016deep,carneiro2015unregistered,massseg}.
However, those methodologies require annotated mass ROI and/or segmentation ground truth.
Dhungel et al. trained a detector and segmentation network on the training set first,
and then used CNN to perform mass classification~\cite{dhungel2016automated}.
The training procedure still requires detection ROI and boundary ground truth, which is costly.
In addition, multi-stage training cannot fully explore the power of the deep network.
Thus, an end-to-end approach for whole mammogram classification is preferred for this problem.

\subsection{Deep Multi-instance Learning}\label{sec:dmil}

Dietterich et al. first proposed multi-instance learning problem~\cite{dietterich1997solving}. There are various other multi-instance related work in the machine learning literature. Andrews et al. generalized support vector machine for the multi-instance problem~\cite{andrews2002multiple}. Kwok and Cheung employed marginalized kernel to solve the instance label ambiguity in multi-instance learning~\cite{kwok2007marginalized}. Zhou et al. extended multi-instance learning to multi-class classification problems~\cite{zhou2012multi}.

Due to the great representation power of deep features~\cite{zhu2016co}, combining multi-instance learning with deep neural networks is an emerging topic. Wu et al. combined CNN with multi-instance learning to auto-annotate natural images~\cite{wu2015deep}. Kotzias et al. incorporated CNN features into multi-instance cost function to do sentiment analysis~\cite{Kotzias:2015:GIL:2783258.2783380}. Yan et al. used a deep multi-instance network to find discriminative patches for body part recognition~\cite{yan2016multi}. Patch based CNN added a new layer after the last layer of deep multi-instance network to learn the fusion model for multi-instance predictions~\cite{hou2015patch}. The above approaches used max pooling to model the general multi-instance assumption which only considered the patch of max probability. In this paper, a more effective task-related deep multi-instance models are explored for whole mammogram classification. 

\section{Deep Multi-instance Networks for Whole Mammogram Mass Classification}\label{sec:dmlmamm}

Leveraging the insights from recent successful deep convolution networks used for natural image processing, we design end-to-end trained deep multi-instance networks for the task. Fig.~\ref{fig:framework} shows the proposed network architecture which has multiple convolutional layers, one linear regression layer, one ranking layer, and one multi-instance loss layer. We employ three schemes for combining multiply instances, 1) the max pooling-based multi-instance learning takes only the largest element from the ranking layer; 2) label assignment-based multi-instance learning utilizes all the elements; and 3) sparse multi-instance learning adds sparse constraints for elements to the ranking layer. The details of these schemes will be detailed later.

The rest of this section is organized as follows. We first briefly introduce the common part of the deep multi-instance networks to make the paper self-contained. Then we introduce the max pooling-based deep multi-instance network in section~\ref{sec:maxpool}. After that, we convert the multi-instance learning into a label assignment problem in section~\ref{sec:labelassign}. Lastly section~\ref{sec:sparse} describes how to inject the priori knowledge that a mass comprises small percentage of a whole mammogram into the deep multi-instance network.

CNN is a successful model to extract deep features from images~\cite{lecun1998gradient}. Unlike other deep multi-instance network~\cite{yan2016multi,hou2015patch}, we use a CNN to efficiently obtain features of all patches (instances) at the same time. Given an image $\bm{I}$, we can get a much smaller feature map $\bm{F}$ of multi-channels $N_c$ after multiple convolutional layers and max pooling layers. The $(\bm{F})_{i,j,:}$ represents deep CNN features for a patch $\bm{Q}_{i,j}$ in $\bm{I}$, where $i,j$ represents the pixel row and column indices respectively, and $:$ denotes the channel dimension.

The goal of our work is to predict whether a whole mammogram contains a malignant mass (BI-RADS $\in \{4, 5, 6\}$ as positive) or not, which is a standard binary class classification problem. We add a logistic regression with weights shared across all the pixel positions following $\bm{F}$. After that, an element-wise sigmoid activation function is applied to the output. The malignant probability of feature space's pixel $(i,j)$ is
\begin{equation}
\label{equ:cnn}
r_{i,j} = \text{sigmoid}(\bm{a} \cdot \bm{F}_{i,j,:} + b),
\end{equation}
where $\bm{a}$ is the weights in logistic regression, and $b$ is the bias, and $\cdot$ is the inner product of the two vectors $\bm{a}$ and $\bm{F}_{i,j,:}$. The $\bm{a}$ and $b$ are shared for different pixel position $i,j$. We can combine $r_{i,j}$ into a matrix $\bm{r} = (r_{i,j})$ of range $[0, 1]$ denoting the probabilities of patches being malignant masses. The $\bm{r}$ can be flattened into a one-dimensional vector as $\bm{r} = (r_{1}, r_{2}, ..., r_{m})$ corresponding to flattened patches $(\bm{Q}_{1}, \bm{Q}_{2}, ..., \bm{Q}_{m})$, where $m$ is the number of patches.

\subsection{Max Pooling-based Multi-instance Learning}\label{sec:maxpool}

The general multi-instance assumption is that if there exists an instance that is positive, the bag is positive. The bag is negative if and only if all instances are negative~\cite{dietterich1997solving}. For whole mammogram classification, the equivalent scenario is that if there exists a malignant mass, the mammogram $\bm{I}$ should be classified as positive. Likewise, negative mammogram $\bm{I}$ should not have any malignant masses. If we treat each patch $\bm{Q}_{i}$ of $\bm{I}$ as an instance, the whole mammogram classification is a standard multi-instance task. 

For negative mammograms, we expect all the $r_i$ to be close to 0. For positive mammograms, at least one $r_i$ should be close to 1. Thus, it is natural to use the maximum component of $\bm{r}$ as the malignant probability of the mammogram $\bm{I}$

\begin{equation}
\label{equ:max}
p(y=1|\bm{I}, \bm{\theta}) = \max\{ r_1, r_2, ..., r_m\},
\end{equation}
where $\bm{\theta}$ is the parameters of deep networks.

If we sort $\bm{r}$ first in descending order as illustrated in Fig.~\ref{fig:framework}, the malignant probability of the whole mammogram $\bm{I}$ is the first element of ranked $\bm{r}$ as
\begin{equation}
\label{equ:sortmax}
\begin{aligned}
&\{{r^\prime}_1, {r^\prime}_2, ..., {r^\prime}_m\} = \text{sort} (\{ r_1, r_2, ..., r_m\}), \\
&p(y=1|\bm{I}, \bm{\theta}) = {r^\prime}_1, \quad\text{and}\quad p(y=0|\bm{I}, \bm{\theta}) = 1-{r^\prime}_1,
\end{aligned}
\end{equation}
where $\bm{r}^\prime = ({r^\prime}_1, {r^\prime}_2, ..., {r^\prime}_m)$ is descending ranked $\bm{r}$. The cross entropy-based cost function can be defined as  
\begin{equation}
\label{equ:maxloss}
\mathcal{L}_{maxpooling} = -\sum_{n=1}^{N} \log(p(y_n | \bm{I}_n, \bm{\theta})) + \frac{\lambda}{2} \|\bm{\theta}\|^2
\end{equation}
where $N$ is the total number of mammograms, $y_n \in \{0,1\}$ is the true label of malignancy for mammogram $\bm{I}_n$, and $\lambda$ is the regularizer that controls model complexity.

Typically, a mammogram dataset is imbalanced, (e.g., the proportion of positive mammograms is about 20\% for the INbreast dataset). In lieu of that, we introduce a weighted loss defined as
\begin{equation}
\label{equ:weightedmaxloss}
\mathcal{L}_{maxpooling} = -\sum_{n=1}^{N} w_{y_n} \log(p(y_n | \bm{I}_n, \bm{\theta})) + \frac{\lambda}{2} \|\bm{\theta}\|^2,
\end{equation}
where $w_{y_n}$ is the empirical estimation of $y_n$ on the training data.

One disadvantage of max pooling-based multi-instance learning is that it only considers the patch ${\bm{Q}^\prime}_1$ (patch of the max malignant probability), and does not exploit information from other patches. A more powerful framework should add task-related priori, such as sparsity of mass in whole mammogram, into the general multi-instance assumption and explore more patches for training. 

\subsection{Label Assignment-based Multi-instance Learning}\label{sec:labelassign}

For the conventional classification tasks, we assign a label to each data point. In the multi-instance learning scheme, if we consider each instance (patch) $\bm{Q}_i$ as a data point for classification, we can convert the multi-instance learning problem into a label assignment problem.

After we rank the malignant probabilities $\bm{r} = (r_{1}, r_{2}, ..., r_{m})$ for all the instances (patches) in a whole mammogram $\bm{I}$ using the first equation in Eq.~\ref{equ:sortmax}, the first few ${r^\prime}_i$ should be consistent with the label of whole mammogram as previously mentioned, while the remaining patches (instances) should be negative. Instead of adopting the general multi-instance learning assumption that only considers the ${\bm{Q}^\prime}_1$ (patch of malignant probability ${r^\prime}_1$), we assume that 1) patches of the first $k$ largest malignant probabilities $\{{r^\prime}_1, {r^\prime}_2, ..., {r^\prime}_k\}$ should be assigned with the same class label as that of whole mammogram, and 2) the rest patches should be labeled as negative in the label assignment-based multi-instance learning.

After the ranking layer using the first equation in Eq.~\ref{equ:sortmax}, we can obtain the malignant probability for each patch
\begin{equation}
\label{equ:ppatch}
\begin{aligned}
p(y=1 | {\bm{Q}^\prime}_i, \bm{\theta}) = {r^\prime}_i, \quad\text{and}\quad p(y=0 | {\bm{Q}^\prime}_i, \bm{\theta}) = 1-{r^\prime}_i.
\end{aligned}
\end{equation}

The weighted cross entropy-based loss function of the label assignment-based multi-instance learning can be defined as

\begin{equation}
\label{equ:weightedlabelloss}
\begin{aligned}
\mathcal{L}_{labelassign.} = &-\sum_{n=1}^{N}  \bigg ( \sum_{j=1}^{k} {w^\prime_{y_n} \log(p(y_n | {\bm{P}^\prime}_j, \bm{\theta}))}\\+ &\sum_{j=k+1}^{m} {w^\prime_{0} \log(p(y=0 | {\bm{P}^\prime}_j, \bm{\theta}))}\bigg )+\frac{\lambda}{2} \|\bm{\theta}\|^2,
\end{aligned}
\end{equation}
where $w^\prime_{y_n}$ is the empirical estimation of $y_n$ based on patch labels
\begin{equation}
\begin{aligned}
w^\prime_{1} = \frac{k \times N_{pos} }{m \times N}, \quad\text{and}\quad w^\prime_{0} = 1 - w^\prime_{1},
\end{aligned}
\end{equation}
where $N_{pos}$ is the number of positive mammograms and $N$ is the total number of mammograms.

One advantage of the label assignment-based multi-instance learning is that it explores all the patches to train the model. Essentially it acts a kind of data augmentation which is an effective technique to train deep networks when the training data is scarce. From the sparsity perspective, the optimization problem of label assignment-based multi-instance learning is exactly a $k$-sparse problem for the positive data points, where we expect $\{{r^\prime}_1, {r^\prime}_2, ..., {r^\prime}_k\}$ being 1 and $\{{r^\prime}_{k+1}, {r^\prime}_{k+2}, ..., {r^\prime}_m\}$ being 0. The disadvantage of label assignment-based multi-instance learning is that it is hard to estimate the hyper-parameter $k$. In our experiment, we choose $k$ based on cross validation. Thus, a relaxed assumption for the multi-instance learning or an adaptive way to estimate the hyper-parameter $k$ is preferred. 

\subsection{Sparse Multi-instance Learning}\label{sec:sparse}

From the mass distribution, the mass typically comprises about 2\% of the whole mammogram on average (Fig.~\ref{fig:mass}), which means the mass region is quite sparse in the whole mammogram. It is straightforward to convert the mass sparsity to the malignant mass sparsity, which implies that $\{{r^\prime}_1, {r^\prime}_2, ..., {r^\prime}_m\}$ is sparse in the whole mammogram classification problem. The sparsity constraint means we expect the malignant probability of part patches ${r^ \prime}_i$ being 0 or close to 0, which is equivalent to the second assumption in the label assignment-based multi-instance learning. Analogously, we expect ${r^\prime}_1$ to be indicative of the true label of mammogram $\bm{I}$.

After the above discussion, the loss function of the sparse multi-instance learning problem can be defined as
\begin{equation}
\label{equ:weightedsparseloss}
\mathcal{L}_{sparse} = \sum_{n=1}^{N} \big ( -w_{y_n}\log(p(y_n | \bm{I}_n, \bm{\theta})) + \mu \|\bm{r}^{\prime}_n\|_1 \big ) +\frac{\lambda}{2} \|\bm{\theta}\|^2,
\end{equation}
where $p(y_n | \bm{I}_n, \bm{\theta})$ can be calculated in Eq.~\ref{equ:sortmax}, $w_{y_n}$ is the same as that in the max pooling based multi-instance learning, $\bm{r}_n = ({r^\prime}_1, {r^\prime}_2, ..., {r^\prime}_m)$ for mammogram $\bm{I}_n$, $\|\cdot\|_1$ denotes the $\mathcal{L}_1$ norm, $\mu$ is the sparsity factor, which is a trade-off between the sparsity assumption and the importance of patch ${\bm{Q}^\prime}_1$.


From the discussion of label assignment-based multi-instance learning, this learning is a kind of exact $k$-sparse problem which can be converted to $\mathcal{L}_1$ constrain. One advantage of sparse multi-instance learning over label assignment-based multi-instance learning is that it does not require assign label for each patch which is hard to do for patches where probabilities are not too large or small. The sparse multi-instance learning considers the overall statistical property of $\bm{r}$. 

Another advantage of sparse multi-instance learning is that, it has different weights for general multi-instance assumption (the first part loss) and label distribution within mammogram (the second part loss), which can be considered as a trade-off between max pooling-based multi-instance learning (slack assumption) and label assignment-based multi-instance learning (hard assumption).

\subsection{Whole Mammogram Classification using the Learned Model}\label{sec:whmamm}

From the above discussion of the three deep multi-instance variants, we always assume the largest probability ${r^\prime}_1$ should be consistent with the malignant label of whole mammogram $\bm{I}$. In the inference, we can take ${p^\prime}_1$ as predicted malignant probability for whole mammogram $\bm{I}$
\begin{equation}
\label{equ:inference}
\begin{aligned}
p(y=1|\bm{I}, \bm{\theta}) = {r^\prime}_1.
\end{aligned}
\end{equation}

\section{Experiments}\label{sec:exp}
We validate the proposed model on the most frequently used mammographic mass classification dataset, INbreast dataset~\cite{moreira2012inbreast}, as the mammograms in other datasets, such as DDSM dataset~\cite{bowyer1996digital} and mini-MIAS dataset~\cite{suckling1994mammographic}, are of low quality. The INbreast dataset contains 410 mammograms of which 94 contains malignant masses. These 94 mammograms with masses are defined as positive mammograms. Five-fold cross validation is used to evaluate model performance. For each testing fold, we use three folds mammograms for training, and one fold for validation to tune the hyper-parameters in the model. The performance is reported as the average of five testing results obtained from the cross-validation.

For preprocessing, we first use Otsu's method to segment the mammogram~\cite{otsu1975threshold} and remove the background of the mammogram. To prepare the mammograms for following CNNs, we resize the processed mammograms to $224 \times 224$. We employ techniques to augment our data. For each training epoch, we randomly flip the mammograms horizontally, shift within 0.1 proportion of mammograms horizontally and vertically, rotate within 45 degree, and set $50 \times 50$ square box as 0. In experiments, the data augmentation is essential for us to train the deep networks.

For the CNN network structure, we use AlexNet and remove the fully connected layers~\cite{krizhevsky2012imagenet}. Through the CNN, the mammogram of size $224 \times 224$ becomes 256 $6 \times 6$ feature maps. Then we use steps in Sec.~\ref{sec:dmlmamm} to do multi-instance learning (MIL). We use Adam optimization with learning rate 0.001 for training from scratch and $5 \times 10^{-5}$ for training models pretrained on the Imagenet \cite{ba2015adam}. The $\lambda$ for max pooling-based and label assignment-based multi-instance learning are $1 \times 10^{-5}$. The $\lambda$ and $\mu$ for sparse multi-instance learning are $5 \times 10^{-6}$ and $1 \times 10^{-5}$ respectively. For the label assignment-based deep multi-instance network, we select $k$ from $\{4,8,12,16\}$ based on the validation set.

We firstly compare our methods to previous models validated on DDSM dataset and INbreast dataset in Table~\ref{tab:inbreast}. Previous hand-crafted feature-based methods required manually annotated detection bounding box or segmentation ground truth~\cite{ball2007digital,varela2006use,domingues2012inbreast}. Pretrained CNN used two CNNs to detect the mass region and segment the mass, followed by a third CNN pretrained by hand-crafted features to do the actual mass classification on the detected ROI region \cite{dhungel2016automated}. Pretrained CNN+RF further used random forest and obtained 7\% improvement. These methods are either manually or semi-automatically, while our methods are totally automated and do not reply on any human designed features or extra annotations.

\begin{table}[t]
	\fontsize{9pt}{10pt}\selectfont\centering
	\caption{Accuracy Comparisons of the proposed deep multi-instance networks and related methods on test sets.}\label{tab:inbreast}
	\begin{tabular}{c|c|c|c|c}
		\hlinew{0.9pt}
		Methodology&Dataset&Set-up&Accu.(\%)&AUC(\%)\\		
		\hlinew{0.7pt}
		\tabincell{c}{Ball et al. \cite{ball2007digital}}&DDSM&Semi-auto.&87&N/A\\
		\hline \tabincell{c}{Varela et al. \cite{varela2006use}}&DDSM&Semi-auto.&81&N/A\\
		\hline \tabincell{c}{Domingues et al. \cite{domingues2012inbreast}}&INbr.&Manual&89&N/A\\
		\hline \tabincell{c}{Pretrained CNN \cite{dhungel2016automated}}&INbr.&Semi-auto.&84$\pm{0.04}$&69$\pm{0.10}$\\
		\hline \tabincell{c}{Pretrained CNN+RF \cite{dhungel2016automated}}&INbr.&Semi-auto.&$\bf{91\pm{0.02}}$&76$\pm{0.23}$\\
		\hlinew{0.9pt}
		AlexNet &INbr.&Auto.&78.30$\pm{0.02}$&66.80$\pm{0.07}$\\
		\hline Pretrained AlexNet &INbr.&Auto.&80.50$\pm{0.03}$&73.30$\pm{0.03}$\\
		\hline \tabincell{c}{AlexNet+Max Pooling MIL} &INbr.&Auto.&83.66$\pm{0.02}$&73.62$\pm{0.05}$\\
		\hline \tabincell{c}{Pretrained AlexNet+Max Pooling MIL} &INbr.&Auto.&86.10$\pm{0.01}$&81.51$\pm{0.05}$\\
		\hline \tabincell{c}{AlexNet+Label Assign. MIL} &INbr.&Auto.&84.16$\pm{0.03}$&76.90$\pm{0.03}$\\
		\hline \tabincell{c}{Pretrained AlexNet+Label Assign. MIL} &INbr.&Auto.&86.35$\pm{0.02}$&82.91$\pm{0.01}$\\
		\hline \tabincell{c}{Pretrained AlexNet+Sparse MIL} &INbr.&Auto.&87.11$\pm{0.03}$&83.45$\pm{0.05}$\\
		\hline \tabincell{c}{Pretrained AlexNet+Sparse MIL+Bagging} &INbr.&Auto.&90.00$\pm{0.02}$&$\bf{85.86\pm{0.03}}$\\
		\hlinew{0.9pt}
	\end{tabular}
\end{table}

From Table ~\ref{tab:inbreast}, we observe the models pretrained on Imagenet, Pretrained AlexNet, Pretrained AlexNet+Max Pooling MIL, and Pretrained AlexNet+Label Assign. MIL, improved 2\%, 3\%, 2\% for AlexNet, max pooling-based deep multi-instance learning (AlexNet+Max Pooling MIL) and label assignment-based deep multi-instance learning (AlexNet+Label Assign. MIL) respectively. This shows the features learned on natural images are helpful for the learning of mammogram related deep network. The label assignment-based deep multi-instance networks trained from scratch obtains better performance than the pretrained CNN using 3 different CNNs and detection/segmentation annotation in the training set. This shows the superiority of our end-to-end deep multi-instance networks for whole mammogram classification. According to the accuracy metric, the sparse deep multi-instance network is better than the label assignment-based multi-instance network, and label assignment-based multi-instance network is better than the max pooling-based multi-instance network. This result is consistent with our previous discussion that the label assignment assumption is more efficient than max pooling assumption and sparsity assumption benefited from not having the hard constraints of the label assignment assumption. We obtained different models by using different validation sets for each test fold and used bagging (voting or average different models' predictions) alleviating overfitting to boost the accuracy. Competitive performance to random forest-based pretrained CNN is achieved.

\begin{figure}[t]
	\begin{center}
		\begin{minipage}{0.49\textwidth}
			\centerline{\includegraphics[width=1.12\textwidth]{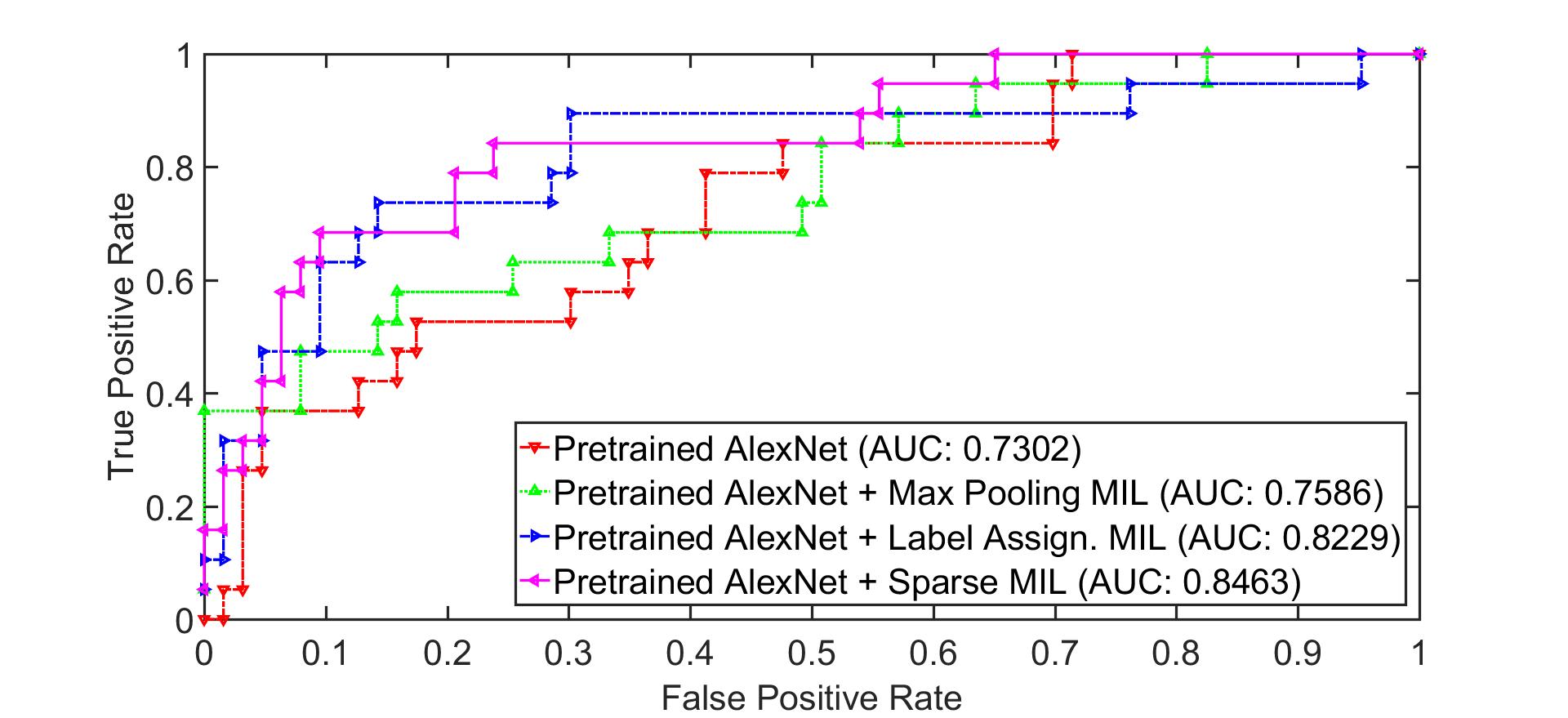}}
			\center{(a)}
		\end{minipage}
		\begin{minipage}{0.49\textwidth}
			\centerline{\includegraphics[width=1.12\textwidth]{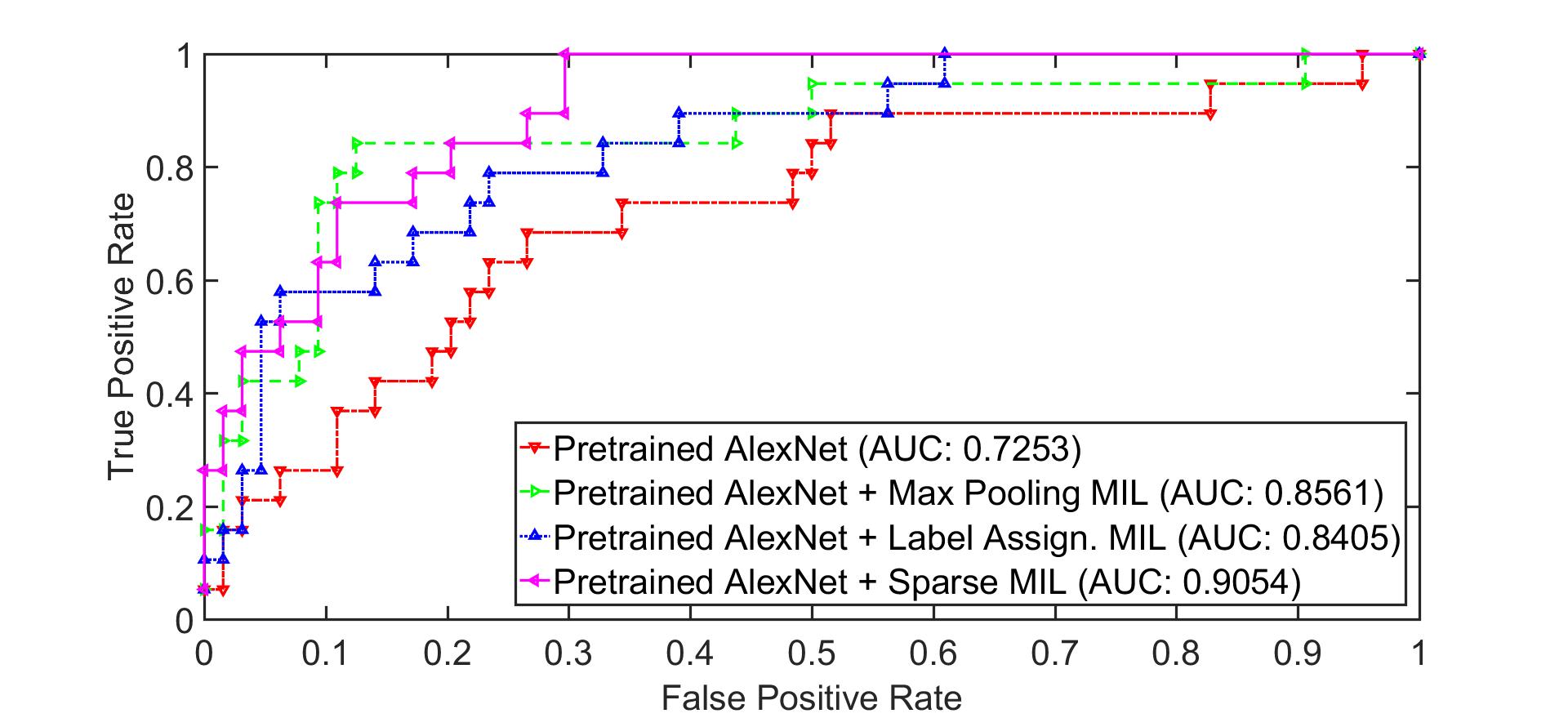}}
			\center{(b)}
		\end{minipage}
		\caption{The ROC curve on fold 2 (a) and fold 4 (b) using pretrained AlexNet, pretrained AlexNet with max pooling multi-instance learning, pretrained AlexNet with label assigned multi-instance learning, pretrained AlexNet with sparse multi-instance learning. The proposed deep multi-instance networks improve greatly over the baseline pretrained AlexNet model.}
		\label{fig:roc}
	\end{center}
\end{figure}

Due to the imbalanced distribution of the dataset where malignant mammograms are only 20\% of total mammograms, the receiver operating characteristic (ROC) curve is a better indicator of performance. We compare the ROC curve on test sets fold 2 and fold 4 in Fig.~\ref{fig:roc} and calculate the averaged area under curve (AUC) of the five test folds in Table ~\ref{tab:inbreast}.


From Fig.~\ref{fig:roc} and Table~\ref{tab:inbreast}, we observe that the sparse deep multi-instance network provides the best AUC, and label assignment-based deep multi-instance network obtains the second best AUC. The deep multi-instance network improves greatly over the baseline models, pretrained AlexNet and AlexNet learned from scratch. The pretraining on Imagenet, Pretrained AlexNet, Pretrained AlexNet+Max Pooling MIL, Pretrained AlexNet+Label Assign. MIL, increases performance of AlexNet, max pooling-based deep multi-instance network (AlexNet+Max Pooling MIL), and label assignment-based deep multi-instance network (AlexNet+Label Assign. MIL) by 7\%, 8\% and 6\%  respectively. This shows the effectiveness and transferability of deep CNN features learned from natural images to medical images. Our deep networks achieves the best AUC result which proves the superior performance of the deep multi-instance networks.

The main reasons for the superior results using our models are as follows. Firstly, data augmentation is an important technique to increase scarce training datasets and proved useful here. Secondly, our models fully explored all the patches to train our deep networks thereby eliminating any possibility of overlooking malignant patches by only considering a subset of patches. This is a distinct advantage over previous networks that employed several stages consisting of detection and segmentation networks. 

\section{Discussions}\label{sec:discus}

\begin{figure}[!t]
	\begin{center}
		\begin{minipage}{0.17\linewidth}
			\centerline{\includegraphics[width=2.5cm]{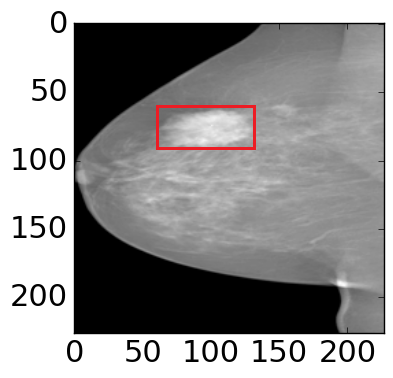}}
			\center{(a)}
		\end{minipage}
		\hspace{1cm}
		\begin{minipage}{0.17\linewidth}
			\centerline{\includegraphics[width=2.5cm]{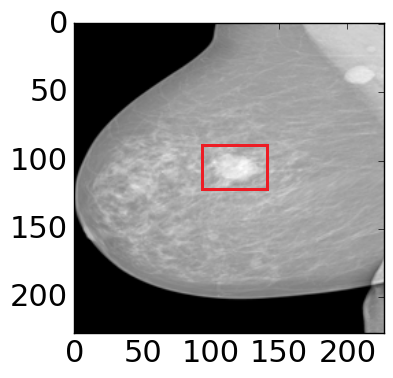}}
			\center{(b)}
		\end{minipage}
		\hspace{1cm}
		\begin{minipage}{0.17\linewidth}
			\centerline{\includegraphics[width=2.5cm]{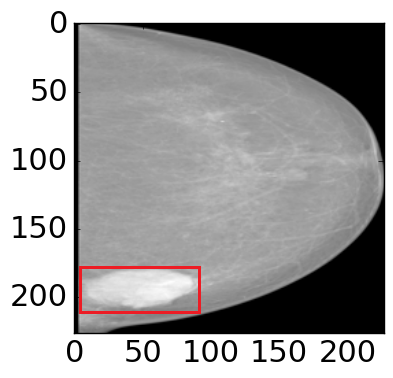}}
			\center{(c)}
		\end{minipage}
		\hspace{1cm}
		\begin{minipage}{0.17\linewidth}
			\centerline{\includegraphics[width=2.5cm]{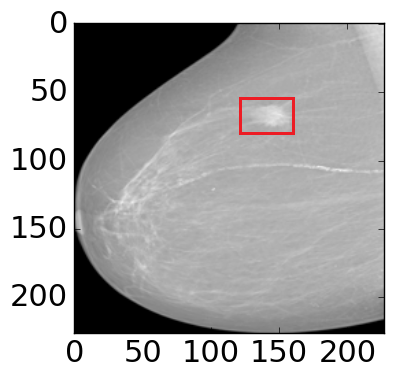}}
			\center{(d)}
		\end{minipage}
		\vfill
		\begin{minipage}{0.17\linewidth}
			\centerline{\includegraphics[width=2.5cm]{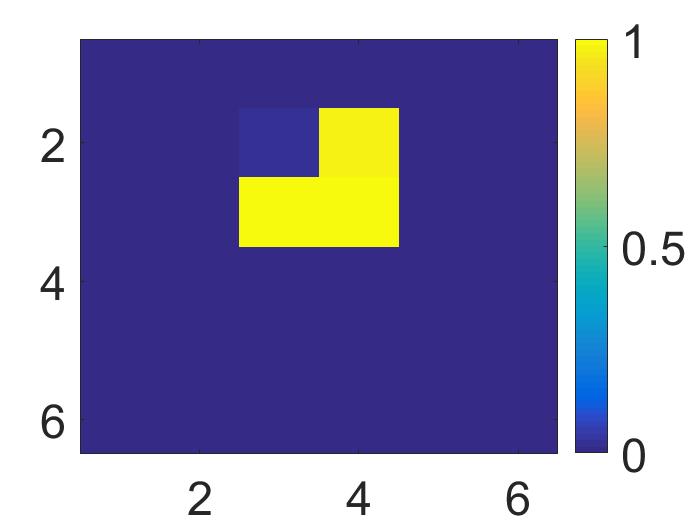}}
		\end{minipage}
		\hspace{1cm}
		\begin{minipage}{0.17\linewidth}
			\centerline{\includegraphics[width=2.5cm]{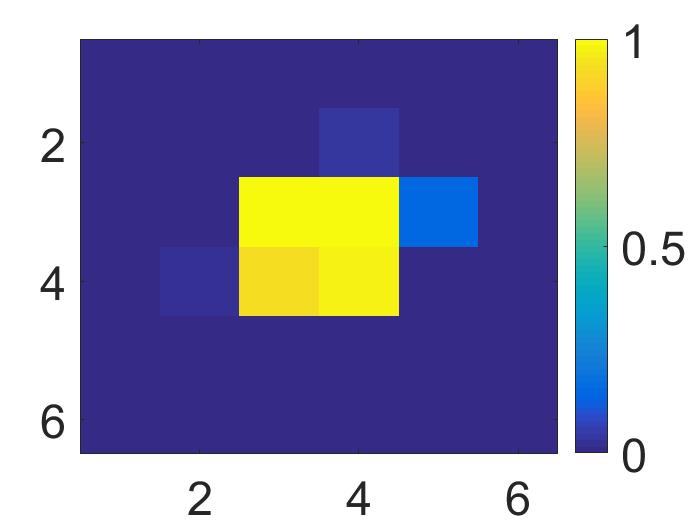}}
		\end{minipage}
		\hspace{1cm}
		\begin{minipage}{0.17\linewidth}
			\centerline{\includegraphics[width=2.5cm]{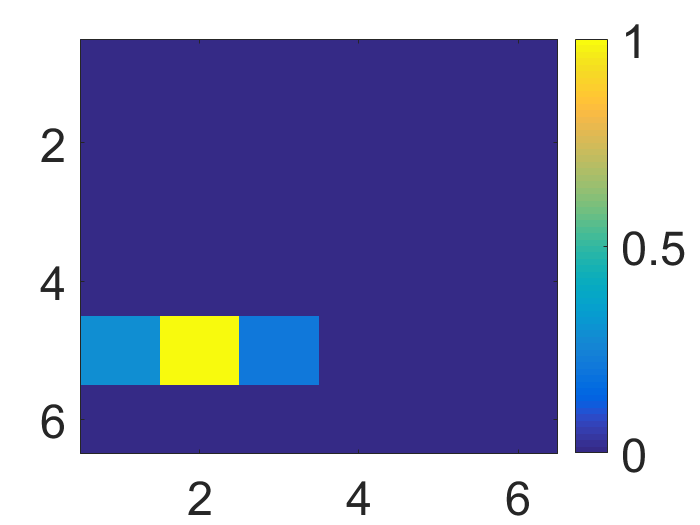}}
		\end{minipage}
		\hspace{1cm}
		\begin{minipage}{0.17\linewidth}
			\centerline{\includegraphics[width=2.5cm]{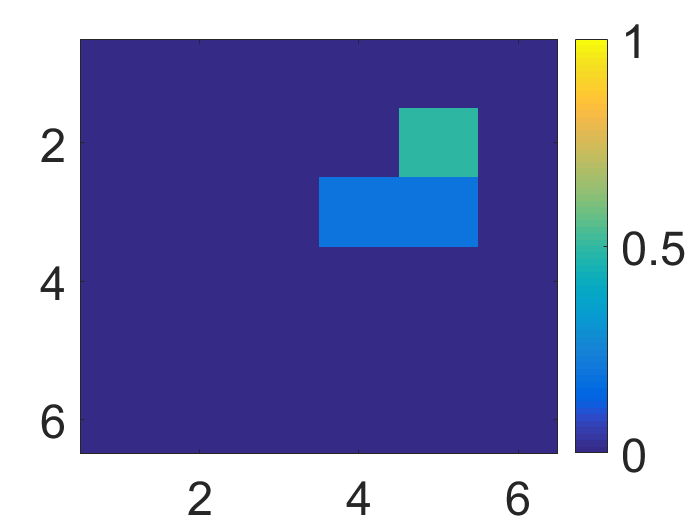}}
		\end{minipage}
		\vfill
		\begin{minipage}{0.17\linewidth}
			\centerline{\includegraphics[width=2.6cm]{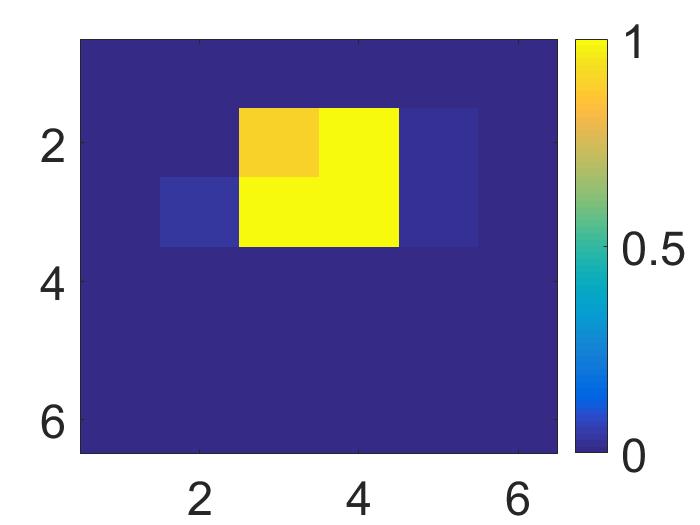}}
		\end{minipage}
		\hspace{1cm}
		\begin{minipage}{0.17\linewidth}
			\centerline{\includegraphics[width=2.6cm]{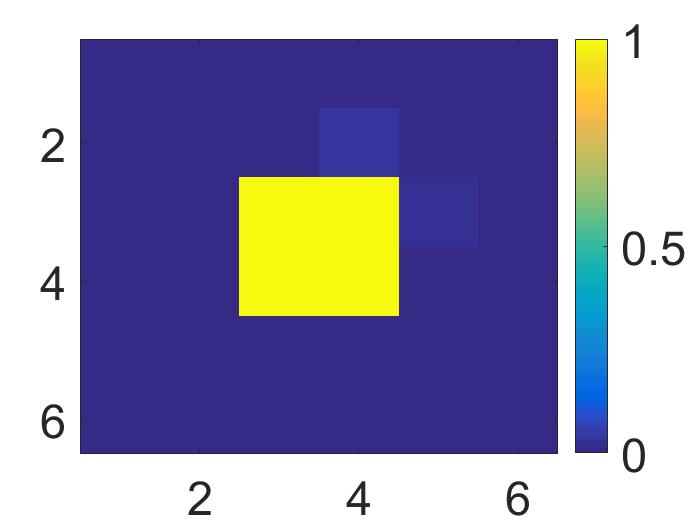}}
		\end{minipage}
		\hspace{1cm}
		\begin{minipage}{0.17\linewidth}
			\centerline{\includegraphics[width=2.6cm]{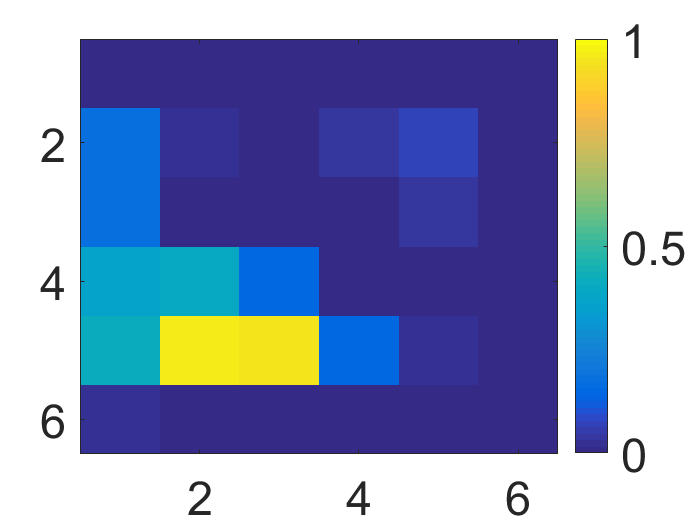}}
		\end{minipage}
		\hspace{1cm}
		\begin{minipage}{0.17\linewidth}
			\centerline{\includegraphics[width=2.6cm]{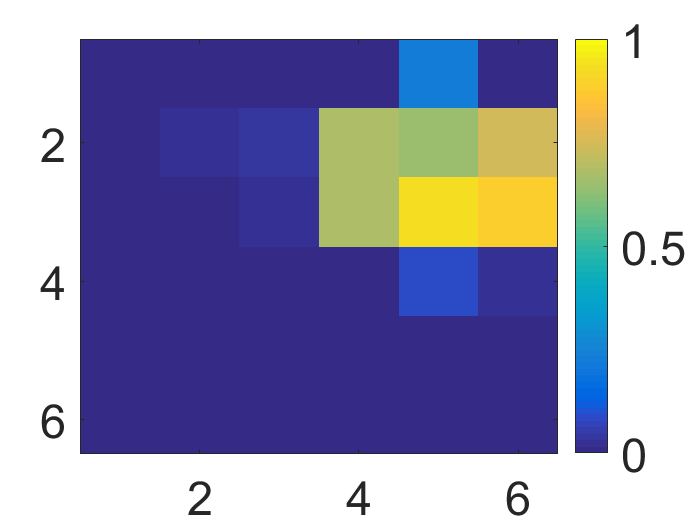}}
		\end{minipage}
		\vfill
		\begin{minipage}{0.17\linewidth}
			\centerline{\includegraphics[width=2.6cm]{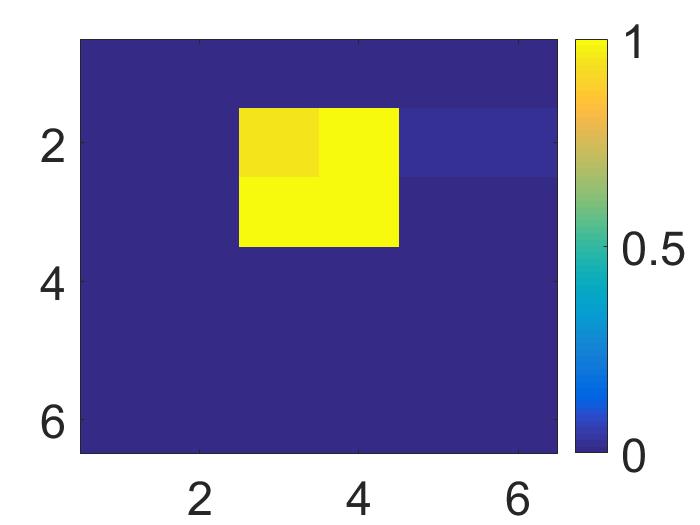}}
		\end{minipage}
		\hspace{1cm}
		\begin{minipage}{0.17\linewidth}
			\centerline{\includegraphics[width=2.6cm]{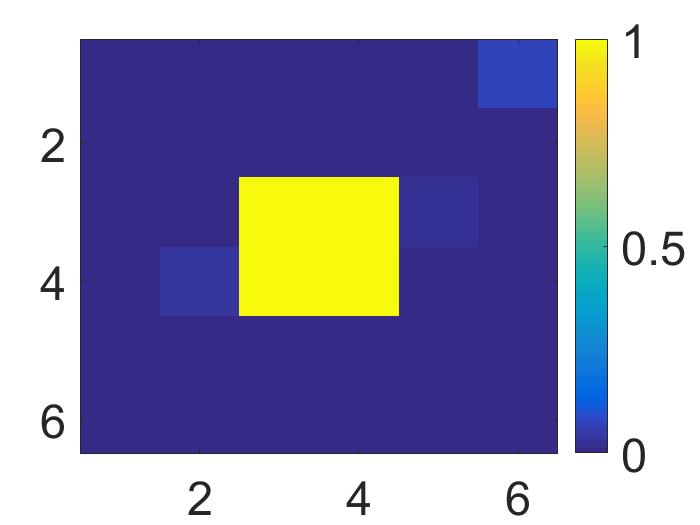}}
		\end{minipage}
		\hspace{1cm}
		\begin{minipage}{0.17\linewidth}
			\centerline{\includegraphics[width=2.6cm]{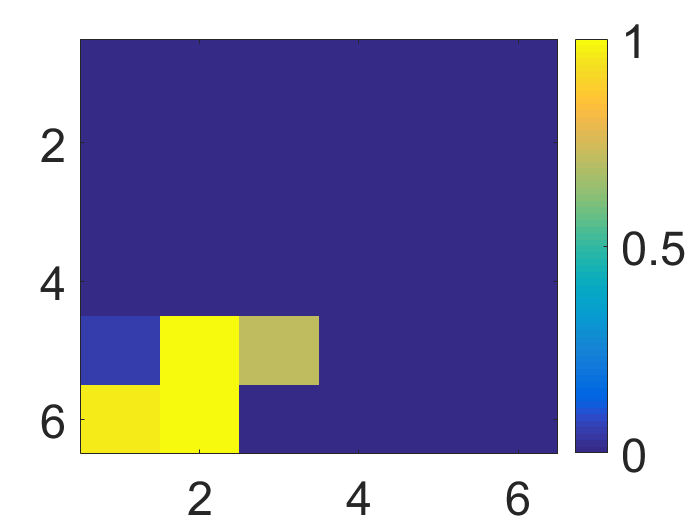}}
		\end{minipage}
		\hspace{1cm}
		\begin{minipage}{0.17\linewidth}
			\centerline{\includegraphics[width=2.6cm]{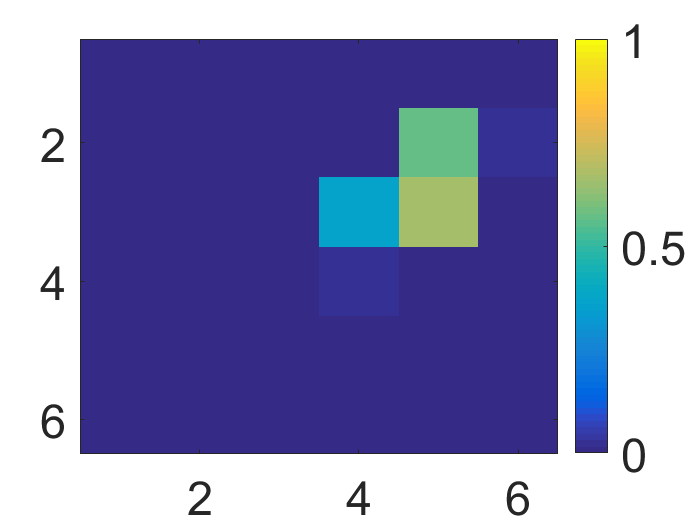}}
		\end{minipage}
		\caption{The visualization of predicted malignant probabilities for instances/patches in four different resized mammograms. The first row is the resized mammogram. The red rectangle boxes are mass regions from the annotations on the dataset. The color images from the second row to the last row are the predicted malignant probability from linear regression layer for (a) to (d) respectively, which are the malignant probabilities of patches/instances. Max pooling-based, label assignment-based, sparse deep multi-instance networks are in the second row, third row, fourth row respectively. Max pooling-based deep multi-instance network misses some malignant patch for mammogram (a), (c) and (d). Label assignment-based deep multi-instance network mis-classifies patches into malignant in (d). }
		\label{fig:visresponse}
	\end{center}
\end{figure}

To further understand our deep multi-instance networks, we visualize the responses of linear regression layer for four mammograms on test set, which represents the malignant probability of each patch, in Fig.~\ref{fig:visresponse}.

From Fig.~\ref{fig:visresponse}, we can see the deep multi-instance network learns not only the prediction of whole mammogram, but also the prediction of malignant patches within the whole mammogram. Our models are able to learn the mass region of the whole mammogram without any explicit bounding box or segmentation ground truth annotation of the training data. The max pooling-based deep multi-instance network misses some malignant patches in (a), (c) and (d). The possible reason is that it only considers the patch of max malignant probability in the training and the model is not well learned for all the patches. The label assignment-based deep multi-instance network mis-classifies some patches in (d). The possible reason is that the model sets a constant $k$ for all the mammograms, which causes some misclassification for small mass. One of the potential applications of our work is that these deep multi-instance learning networks could be used to do weak mass annotation automatically, which is important for computer-aided diagnosis. 

\section{Conclusion}\label{sec:con}

In this paper, we proposed end-to-end trained deep multi-instance networks for whole mammogram classification. Different from previous works using segmentation or detection annotations, we conducted mass classification based on whole mammogram directly. We convert the general multi-instance learning assumption to label assignment problem after ranking. Due to the sparsity of masses, sparse multi-instance learning is used for whole mammogram classification. We explore three schemes of deep multi-instance networks for whole mammogram classification. Experimental results demonstrate more robust performance than previous work even without detection or segmentation annotation in the training.

In future works, it is promising to extend the current work by: 1) incorporating multi-scale modeling such as spatial pyramid to further improve whole mammogram classification, 2) adaptively estimating the parameter $k$ in the label assignment-based multi-instance learning, and 3) employing the deep multi-instance learning to do annotation or provide potential malignant patches to assist diagnoses. Our method should be generally applicable to other bio-image analysis problems where domain expert knowledge and manual labeling required, or region of interest is small and/or sparse relative to the whole image. Our end-to-end deep multi-instance networks are also suited for the large datasets and expected to have improvement if the big dataset is available.



\bibliographystyle{splncs03}

\bibliography{typeinst}

\end{document}